\documentclass[sigconf]{acmart}

\usepackage{multirow}
\usepackage{float}
\usepackage{siunitx}
\usepackage{balance}
\usepackage[utf8]{inputenc}
\usepackage{xcolor}
\usepackage{acmart-taps}

\newcolumntype{j}{S[table-format=2.2]}

\copyrightyear{2023} 
\acmYear{2023} 
\setcopyright{acmlicensed}\acmConference[SOICT 2023]{The 12th International Symposium on Information and Communication Technology}{December 7--8, 2023}{Ho Chi Minh, Vietnam}
\acmBooktitle{The 12th International Symposium on Information and Communication Technology (SOICT 2023), December 7--8, 2023, Ho Chi Minh, Vietnam}
\acmPrice{15.00}
\acmDOI{10.1145/3628797.3628921}
\acmISBN{979-8-4007-0891-6/23/12}

\begin{document}

\title{Abusive Span Detection for Vietnamese Narrative Texts}

\author{Nhu-Thanh Nguyen}
\affiliation{%
  \institution{University of Information Technology}
  \country{Ho Chi Minh City, Vietnam}
}
\affiliation{%
  \institution{Vietnam National University}
  \country{Ho Chi Minh City, Vietnam}
}
\email{thanhnn.15@grad.uit.edu.vn}

\author{Khoa Thi-Kim Phan}
\affiliation{%
  \institution{University of Information Technology}
  \country{Ho Chi Minh City, Vietnam}
}
\affiliation{%
  \institution{Vietnam National University}
  \country{Ho Chi Minh City, Vietnam}
}
\email{khoaptk@uit.edu.vn}

\author{Duc-Vu Nguyen}
\affiliation{%
  \institution{University of Information Technology}
  \country{Ho Chi Minh City, Vietnam}
}
\affiliation{%
  \institution{Vietnam National University}
  \country{Ho Chi Minh City, Vietnam}
}
\email{vund@uit.edu.vn}

\author{Ngan Luu-Thuy Nguyen}
\affiliation{%
  \institution{University of Information Technology}
  \country{Ho Chi Minh City, Vietnam}
}
\affiliation{%
  \institution{Vietnam National University}
  \country{Ho Chi Minh City, Vietnam}
}
\email{ngannlt@uit.edu.vn}

\begin{abstract}
Abuse in its various forms, including physical, psychological, verbal, sexual, financial, and cultural, has a negative impact on mental health. However, there are limited studies on applying natural language processing (NLP) in this field in Vietnam. Therefore, we aim to contribute by building a human-annotated Vietnamese dataset for detecting abusive content in Vietnamese narrative texts. We sourced these texts from VnExpress, Vietnam's popular online newspaper, where readers often share stories containing abusive content. Identifying and categorizing abusive spans in these texts posed significant challenges during dataset creation, but it also motivated our research. We experimented with lightweight baseline models by freezing PhoBERT and XLM-RoBERTa and using their hidden states in a BiLSTM to assess the complexity of the dataset. According to our experimental results, PhoBERT outperforms other models in both labeled and unlabeled abusive span detection tasks. These results indicate that it has the potential for future improvements.\end{abstract}

\ccsdesc[500]{Computing methodologies~Natural language processing}

\keywords{Abusive Span Detection, Vietnamese Narrative Texts Dataset, Pre-trained Language Models, Long Short-Term Memory, Sequence Labeling}

\maketitle

\section{Introduction}
\sloppy Abuse is common worldwide and has substantial adverse consequences \cite{radell2021impact}. With the main purpose of maintaining power and control over another, abuse can be categorized into some prevalent forms, such as Physical, Sexual, Financial/Economic, Cultural/Identity, Verbal/Emotional, and Mental/Psychological \cite{reachma}. Understanding its influences, and severity, many works have been promoted, involving academic fundamental research, and practical applications \cite{shepard1992abusive, yon2019prevalence, emezue2020digital, radell2021impact}. One of these practical applications is leveraging the proliferation of social media to create a trustworthy environment, consisting of confession forums, and sharing sections for abused people to freely express their problems. Therefore, the amount of narrative text data containing abusive phenomena is abundant. 

\begin{figure}[H]
    \centering
    \includegraphics[width=\columnwidth]{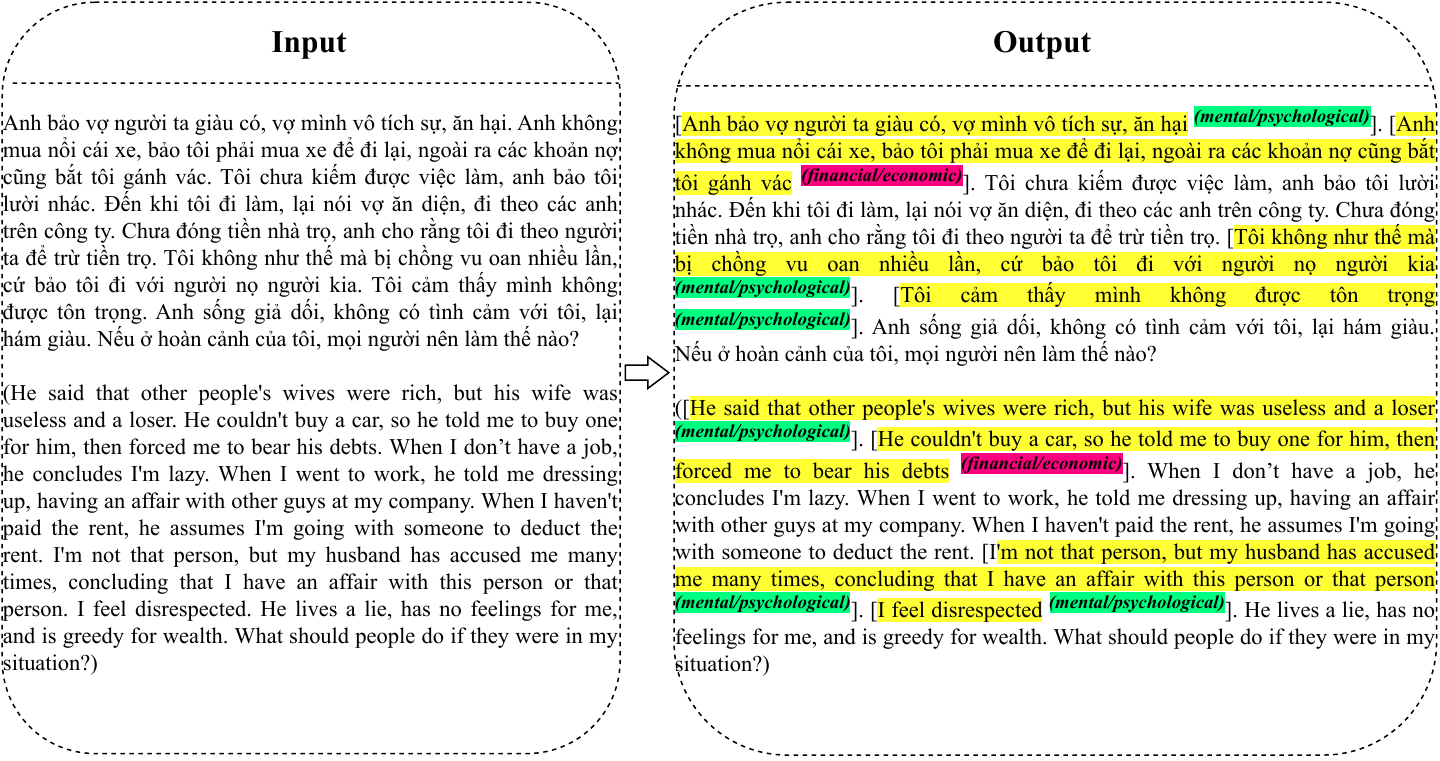}
    \caption{An example for Abusive span detection for Vietnamese narrative texts.}
    \label{fig1}
\end{figure}

In recent years, many works have used Natural Language Processing techniques to automatically extract important information from social media resources. This facilitates meaningful applications such as chatbots, question-answering, and recommendation systems.
However, before mining the data, the exact classification of narrative text data is not easy in this age. Most previous studies focus on research done using social media data. In particular, tweets, and comments that are short and lack a sequence of events are preferred to exploit, especially in the field of detecting abuse \cite{al2022natural, davidson2017automated,kumar-etal-2018-benchmarking, emon2022detection}.

Similarly, in the Vietnamese domain, there are also some related works about hate and offensive language mainly mining social media data such as comments, including ViHOS \cite{hoang-etal-2023-vihos}, ViHSD \cite{luu2021large}, HSD-VLSP \cite{vu2020hsd}, and UIT-ViCTSD \cite{nguyen2021constructive}. Although the definitions of abusive and offensive language are closely related, there are still important distinctions to define. According to \cite{caselli-etal-2020-feel}, while both are associated with ``strongly impolite, rude``, the use of abusive language often involves an intention. To our knowledge, the monolingual dataset for recognizing abusive spans in Vietnamese narrative text is rare and may not exist. This questions us where we find a reliable source, and how we can address the task of recognizing abusive spans in Vietnamese narrative texts, which can promote many useful applications in the health science field. Therefore, we would like to (i) create a new dataset for the task of abusive spans detection in Vietnamese truly narrative texts as depicted in Figure \ref{fig1} to not only the Natural Language Processing research community but also the health science field in the future, (ii) develop a new task, namely abusive spans detection described as follows:
    \begin{itemize}
        \item \textbf{Input}: a Vietnamese narrative text \textbf{\textit{P}} that consists of \textbf{N} characters.
        \item \textbf{Output}: One or more spans describing abuse are detected directly from the narrative text \textbf{\textit{P}} for each category. Each span is extracted from the position \textit{i} to position \textit{j} such that $0 \leq \textit{i}$, $\textit{j} \leq \textbf{N}$, and $\textit{i} \leq \textit{j}$. 
    \end{itemize}

In this work, we have two main contributions summarized as follows:
\begin{itemize}
    \item We created the first human-annotated dataset for Abusive spans detection for Vietnamese narrative texts, comprising 519 spans in 198 abusive texts in 1041 narrative texts annotated with 6 aspect categories: Physical, Sexual, Financial/Economic, Cultural/Identity, Verbal/Emotional, and Mental/Psychological. To build a benchmark narrative text dataset, we require a reliable source. Fortunately, VnExpress which is a well-known Vietnamese online newspaper has a narrative section. The VnExpress's narrative section \cite{vnexpress} contains personal narrative stories submitted by readers and standardized by editors. We crawled all stories which were posted in this section as a source to build our dataset. Besides, our dataset is annotated with a clear definition of abuse categories and detailed guidelines based on the study of REACH community \cite{reachma}.
    \item To evaluate the challenge of the dataset to models, we conducted experiments with lightweight baseline models by freezing PhoBERT \cite{phobert}, XLM-RoBERTa \cite{xlmroberta}, and using their hidden states in a BiLSTM \cite{hochreiter1997long}, combining with two kinds of decoders for the classifier layer: Softmax and CRF \cite{sutton2012introduction}. We also investigated the dataset in two tasks: unlabeled abusive span detection, and labeled abusive span detection. We attained (i) the relaxed evaluation inspired by the approach in \cite{hoang-etal-2023-vihos} brings better results than the strict one which requires precise prediction of the start and end points of a span; (ii) using CRF as the final decoder layer provides better predictions than the Softmax layer; (iii) using {$\text{PhoBERT}_\text{large}$}  combined with BiLSTM-CRF gives the highest overall results, at 86.00\% and 58.10\% respectively for the relaxed evaluation for both two tasks.
\end{itemize}

In Section 2, we review prior research on abusive language detection, including approaches and datasets from Vietnamese and other domains. Sections 3 and 4 cover our dataset creation and analysis. In Section 5, we detail our approach to the new task, and Section 6 concludes and outlines future work.

\section{Related Work} 
In recent years, abusive language detection and other related problems such as offensive language have attracted much attention from the Natural Language Processing (NLP) community. Research in the field has largely considered some particular topics such as HateSpeech \cite{davidson2017automated, pavlopoulos-etal-2021-semeval}, Cyberbullying \cite{ali2018cyberbullying, afrifa2022cyberbullying, emon2022detection, al2022natural}, and Sexism/Racism \cite{waseem2016hateful, chiril2019multilingual, jiang2022swsr}. By and large, these works have been conducted in English or other resource-rich languages, including Chinese, Spain, and French. Besides, the majority of datasets for these tasks have been collected from two main popular social media platforms these days, consisting of Twitter \cite{davidson2017automated, al2022natural}, and Facebook \cite{kumar-etal-2018-benchmarking, emon2022detection}. Similarly, in the Vietnamese domain, most of the works related to the field have also mainly focused on Hate Speech such as ViHSD \cite{luu2021large}, HSD-VLSP \cite{vu2020hsd}, UIT-ViCTSD \cite{nguyen2021constructive}, and Vi-HOS \cite{hoang-etal-2023-vihos} based on comments crawling from social media platforms. All these datasets represent multi-class classification problems, except the SemEval-2021 Task5: Toxic Spans Detection dataset \cite{pavlopoulos-etal-2021-semeval} for English and the Vi-HOS: Hate Speech Spans Detection \cite{hoang-etal-2023-vihos} for Vietnamese, which entail multi-label classification. 

Most of the works related to the field of detecting abusive, offensive, and hate speech have employed traditional machine learning models such as character n-gram Logistic Regression \cite{grondahl2018all}, and Support Vector Machines \cite{pamungkas2020misogyny}  and models involving deep neural networks combined with word embeddings \cite{zhang2018detecting, ranasinghe2019emoji, arango2022hate}. Since the emergence of BERT \cite{devlin2019bert}, more works have been established based on itself and its variants, achieving competitive results in shared tasks such as SemEval-2019 task 6 \cite{zampieri2019semeval}, SemEval-2020 task 12 \cite{zampieri2020semeval}, and SemEval-2021 task 5 \cite{pavlopoulos2021semeval}. For the task of span detection, there are a few works in English such as the teams submitting to the shared task named SemEval-2021 Task 5: Toxic Spans Detection \cite{pavlopoulos-etal-2021-semeval} and the HateXplain dataset \cite{mathew2021hatexplain}, and in Vietnam consisting of ViHOS: Hate Speech Spans Detection for Vietnamese \cite{hoang-etal-2023-vihos}. In general, the above works used a mixture of transformer-based models and neural network methods. In particular, the best-performing team (HITSZ-HLT) in SemEval-2021 Task 5 \cite{zhu2021hitsz} employed two systems based on BERT \cite{devlin2019bert}, in which for one system, a Conditional Random Field (CRF) layer \cite{sutton2012introduction} was added on top, and for another, an LSTM layer \cite{graves2012long} was added between BERT and CRF layer. For the HateXplain dataset \cite{arango2022hate}, they investigated several models such as CNN-GRU, BiRNN, BiRNN-Attention, and BERT on their dataset. In the ViHOS \cite{hoang-etal-2023-vihos}, they used strong baseline models such as BiLSTM-CRF \cite{bilstm-crf}, XLM-RoBERTa \cite{xlmroberta}, and PhoBERT \cite{phobert} to evaluate the efficacy of their dataset.

Besides these above works, we argue that abusive language detection in NLP needs to be expanded to details by recognizing general types of abuse, particularly 6 types of abuse according to the study of REACH community \cite{reachma}: Physical, Sexual, Financial/Economic, Cultural/Identity, Verbal/Emotional, and Mental/Psychological, namely Abusive spans detection. These categories of abuse have been noted in the health science field \cite{shepard1992abusive, radell2021impact, reachma}, but have not yet received the same level of attention in NLP. To our knowledge, there is still little research on Abusive span detection in narrative texts worldwide, as well as in Vietnamese. Therefore, in this study, we would like to develop\textbf{ the first Vietnamese benchmark dataset for the new task namely Abusive spans detection in Vietnamese narrative texts}. 

\section{Dataset Creation}
Inspired by \cite{annotationbestpratice}, we defined the process with four phases: data collection, guideline creation, training, and annotation.
\begin{figure}[H]
    \centering
    \includegraphics[width=0.9\linewidth]{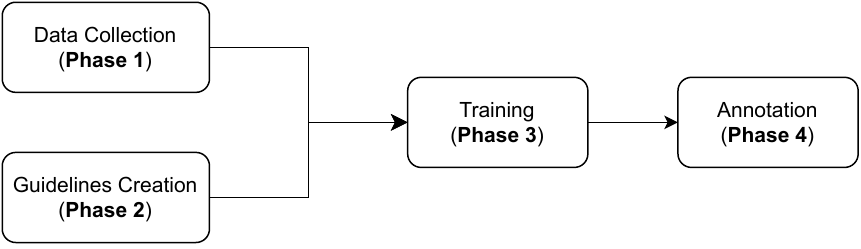}
    \caption{Overview dataset creation process}
    \label{fig:enter-label-1}

\end{figure}

\subsection{Phase 1 - Data Collection}
We crawled 4,800 stories from VnExpress's narrative section \cite{vnexpress}. Each story has been stored as a .txt file in our raw dataset. After that, we randomly selected 1,041 texts from the 4,800 crawled texts to annotate and experiment. The remaining texts in our raw dataset will be annotated in the future.

\subsection{Phase 2 - Guidelines Creation}
\begin{figure}[H]
    \centering
    \includegraphics[width=0.75\linewidth]{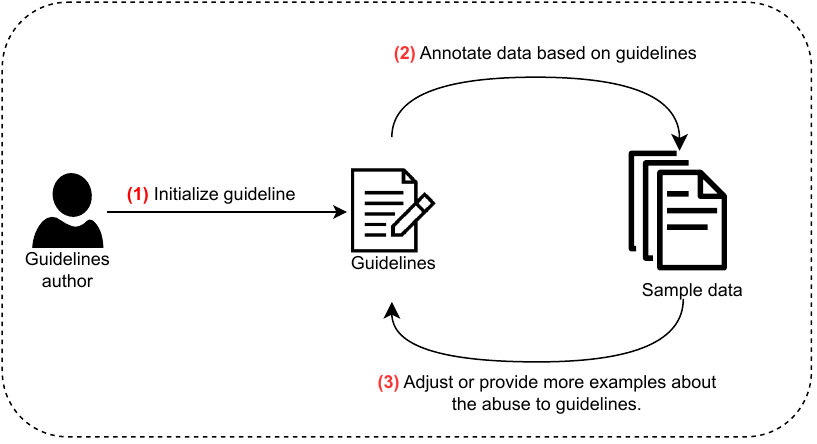}
    \caption{Guidelines Creation phase}
    \label{fig:enter-label-2}

\end{figure}
The objective of this stage is to provide clear instructions for annotators to ensure they can annotate data properly. The guidelines focus on identifying and categorizing abusive segments within texts. We have established a two-step process, which includes the following:\\ 
\indent \textbf{Step 1:} For each sentence, determine if it contains abusive content. To do that, annotators will be based on the definition of abuse, along with a list of common words or phrases that are frequently used. Although we have a list of common words or phrases, annotators also need to read the contents of the sentence to determine.\\
\indent \textbf{Step 2:} Categorize the type of abuse if the sentence contains abuse. Annotators will based on the definition of different types of abuse. Each type of abuse relates to specific aspects, for example:
\begin{itemize}
    \item \textit{Physical} abuse can involve punching, hitting, slapping, kicking, ... a partner against their will.
    \item \textit{Verbal/Emotional} abuse is when someone uses words to criticize, insult, or harm the emotional well-being of others.
    \item \textit{Cultural/Identity} abuse is when abusers use aspects of a victim's cultural identity to inflict suffering or control them. For example, threatening to expose someone's LGBQ/T identity if their loved ones do not know.
\end{itemize}
\begin{itemize}
    \item \textit{Sexual} abuse can include both physical and non-physical components. It can involve rape or other forced sexual acts, or withholding or using sex as a weapon. An abusive partner might also use sex as a means to judge their partner and assign a value.
\end{itemize}
\begin{itemize}
    \item \textit{Financial/Economic} abuse is when abusers use finances to maintain the power and control. Whether it is controlling all of the budgeting in the household and not letting the survivor have access to their own bank accounts or spending money, or opening credit cards and running up debts in the survivor’s name.
\end{itemize}
\begin{itemize}
    \item \textit{Mental/Psychological} abuse happens when one partner, through a series of actions or words, wears away at the other’s sense of mental well-being and health.
\end{itemize}
Besides that, we also provided situational examples for each abuse type to aid annotators' understanding.

\subsection{Phase 3 - Training Annotators}
In this phase, we train annotators to ensure they understand the guidelines well. We also let them practice labeling on a sample set, including abusive and non-abusive content. 
After the annotators finished their tasks, the guidelines were reviewed and revised. Any inconsistencies or divergences were resolved based on guidelines. Then, the difference between the first results and the final results were clarified with the annotators. In this phase, the guidelines may be adjusted and enhanced if required.
\begin{figure} [H]
    \centering
    \includegraphics[width=1\linewidth]{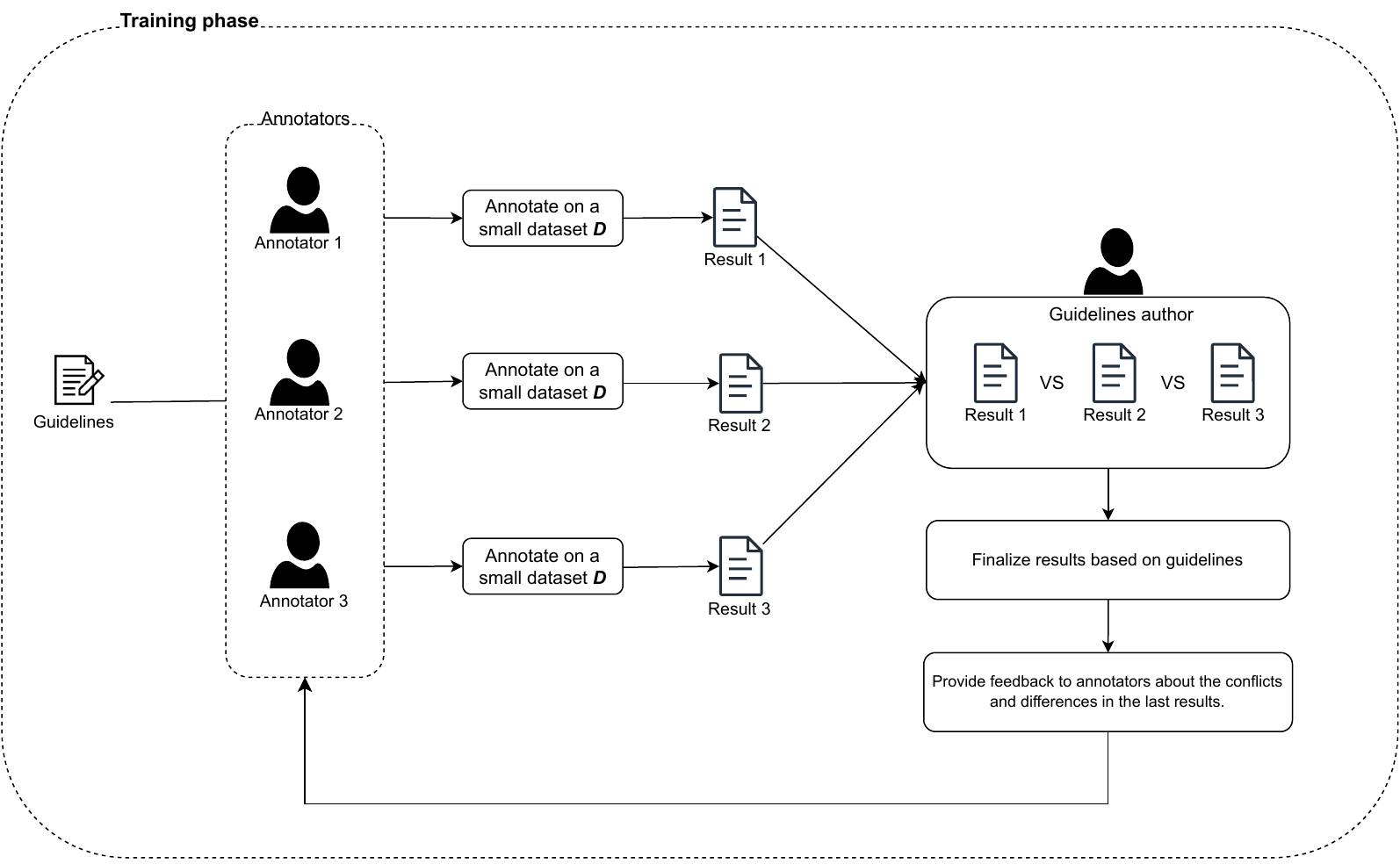}
\caption{Training phase}
    \label{fig:enter-label-3}
\end{figure}

After the training, we measured the agreement of our three annotators to ensure they were on the same page. We randomly selected a set of 50 abusive texts. And each annotator will annotate independently. Then, we collected the results and evaluated them based on F-score. For the first round, the F-score was 76.09\%, and after re-training, the F-score increased to 80.12\%

\subsection{Phase 4 - Annotation}
Each annotator was assigned a specific set of texts to label using the sequence labeling method. If there was any uncertainty regarding to the label, annotators would have a discussion to decide on the final label. Following this stage, we divided our dataset into train, dev, and test sets for experimentation.

\section{Dataset Analysis}
\begin{table}[H]
\centering
\resizebox{\columnwidth}{!}{%
\begin{tabular}{lrrrr}
\hline
    & \textbf{Train} & \textbf{Dev} & \textbf{Test} & \textbf{All}\\\hline
    Number of texts & 700 & 100 & 241 & 1041 \\ 
    Number of abusive texts & 133 & 21 & 44 & 198 \\ 
    Number of non-abusive texts & 567 & 79 & 197 & 843 \\
    Total unlabeled-abusive span count & 350 & 45 & 118 & 513 \\
    Total labeled-abusive span count & 355 & 45 & 119 & 519 \\ \hline
\end{tabular}
}
\caption{Overall statistics of the entire dataset}
\label{tab: overall-stat}
\vspace{-4mm}
\end{table}
Our dataset consists of 1041 narrative texts, which we divided into three sets: 700 for the train set, 100 for the dev set, and 241 for the test set.
Our study explored two tasks: \textbf{(1)} unlabeled abusive span detection, and \textbf{(2)} labeled abusive span detection.
\begin{itemize}
    \item For task \textbf{(1)}, abusive spans in our dataset have only one label: \textit{abusive}.
    \item For task \textbf{(2)}, abusive spans in our dataset have one in six labels representing six abuse types.
\end{itemize}

\begin{figure}[H]
    \centering
    \includegraphics[width=1\linewidth]{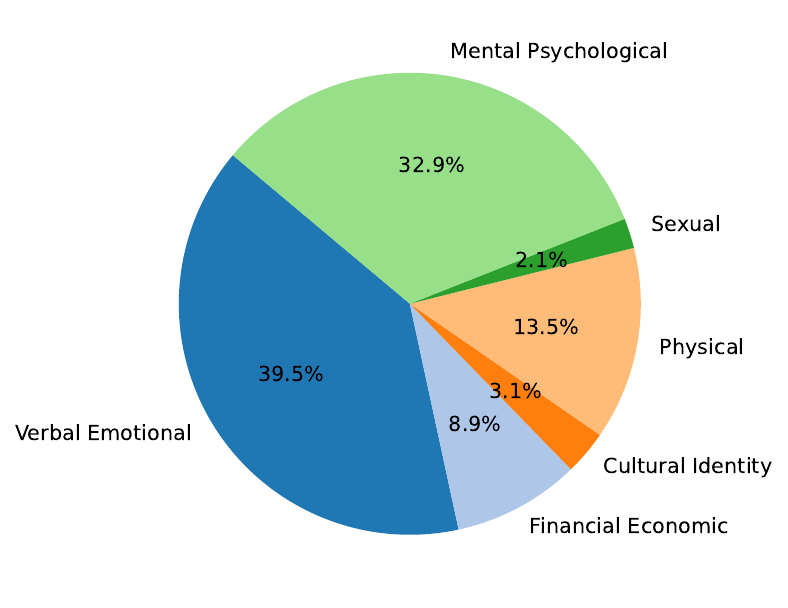}
    \caption{Distribution of abusive types across the entire dataset}
    \label{fig:overall-distribution}
\end{figure}

Before conducting the experiments, we analyzed the dataset to understand it better and facilitate evaluating the results. Table \ref{tab: overall-stat} provides the overall statistics of the entire dataset.
As shown in this table, abusive texts make up 19.02\% of the entire dataset. It indicates that narrative texts containing abuse are lower than non-abusive narrative texts. Consequently, the accuracy of predicting abuse may be impacted. This is one of the challenges, which we will see through the experiment results.

\begin{figure}[!ht]
    \centering
    \includegraphics[width=1\linewidth]{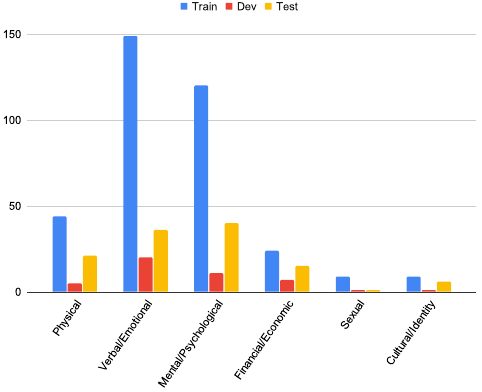}
    \caption{Distribution of abusive types across Train, Dev, and Test sets}
    \label{fig:distribution-3set}
    \vspace{-2mm}
\end{figure}

As we explored the classification of abusive behavior in our dataset, we observed an imbalance among different abusive types. The distribution of abusive types throughout the entire dataset can be observed in Figure \ref{fig:overall-distribution}, while Figure \ref{fig:distribution-3set} illustrates the distribution of abusive types in the Train, Dev, and Test sets individually.

The charts show that \textit{Verbal/Emotional} and \textit{Mental/Psychological} makeup comprises over 60\% of the data, while the distribution of \textit{Sexual} (2.1\% in total) and \textit{Cultural/Identity} (3.1\% in total) types are significantly lower. The lack of data on \textit{Sexual} and \textit{Cultural/Identity} abuse types may potentially affect the predicted results for these types.

Furthermore, when examining the average length of an abusive span, we observe that the average length of a span in our dataset is quite high. In Table \ref{tab:average-words}, the longest average length in our dataset is 30.44 words (\textit{Financial/Economic} type). And our unlabeled dataset version has an average of 26.59 words per span. The length of abusive spans is a factor that makes our dataset more challenging. Besides that, to evaluate the results, we employed strict and relaxed methods. Because the abusive span is relatively long, which leads to the strict evaluation provided low results.

\begin{table}[ht]
    \centering
    \begin{tabular}{lr}
    \hline
         & \textbf{Avg span's length}\\\hline
        Physical & 22.43 \\ 
        Verbal/Emotional & 28.67 \\ 
        Mental/Psychological & 24.99 \\
        Financial/Economic & 30.44 \\
        Sexual & 26.38 \\ \hline
    \end{tabular}
    \caption{Average  span's length of each abusive type}
\label{tab:average-words}
\vspace{-7mm}
\end{table}

\section{Empirical Evaluation}

\subsection{Baseline Models}
To evaluate the complexity of our dataset, we experimented with some effective baseline models for the Vietnamese language. Our approaches are freezing PhoBERT and XLM-RoBERTa and using their hidden states in a BiLSTM. For the classifier layer, we utilized two kinds of decoders: Softmax and CRF.

\begin{table}[ht]
\centering
\begin{tabular}{l|l}
\hline
\textbf{Pre-trained Language Model}              & \textbf{RNN-Decoder layer} \\ \hline
\multirow{2}{*}{$\text{PhoBERT}_\text{base}$}  & BiLSTM-Softmax    \\ \cline{2-2} 
                               & BiLSTM-CRF        \\ \hline
\multirow{2}{*}{$\text{PhoBERT}_\text{large}$} & BiLSTM-Softmax    \\ \cline{2-2} 
                               & BiLSTM-CRF    \\ \hline
\multirow{2}{*}{$\text{XLM-R}_\text{base}$}    & BiLSTM-Softmax    \\ \cline{2-2} 
                               & BiLSTM-CRF        \\ \hline
\multirow{2}{*}{$\text{XLM-R}_\text{large}$}   & BiLSTM-Softmax    \\ \cline{2-2} 
                               & BiLSTM-CRF        \\ \hline
\end{tabular}
\caption{All baseline models were used in our experiments}
\label{tab:exp-summary}
\vspace{-6mm}
\end{table}

Table \ref{tab:exp-summary} shows all baseline models we used in our experiments:
\begin{itemize}
    \item \textbf{BiLSTM} \cite{hochreiter1997long}: a recurrent neural network that can process the input in both directions and utilize information from both sides.
    \item \textbf{BiLSTM-CRF} \cite{bilstm-crf}: a BiLSTM combines with Conditional Random Field (CRF) \cite{laffertyCrf} layer. The BiLSTM is used to process each sentence token-by-token, and then its results are used as input for CRF to produce the prediction results. This model achieved the highest performance in span detection tasks \cite{pavlopoulos-etal-2021-semeval}.
    \item \textbf{XLM-RoBERTa} \cite{xlmroberta}: a multilingual language model that supports 100 languages. It was trained on 2.5 TB of newly created clean CommonCrawl data, including 137 GB of Vietnamese text.
    \item \textbf{PhoBERT} \cite{phobert}: a monolingual language model pre-trained on a 20 GB Vietnamese dataset that performs the most efficiently for the Vietnamese language. Its approach is based on RoBERTa \cite{liu2019roberta}, which optimizes the BERT \cite{devlin2019bert} for more robust performance.
\end{itemize}

\subsection{Experimental Settings}
Regarding all pre-trained language models, the frozen hidden states were projected to a dimension of 128, which was then used as the input for the BiLSTM. To facilitate a lightweight experiment, the BiLSTM's hidden size was configured at 64, while both the word embedding dropout and recurrent dropout were set to 0.1. Instead of using the entire word as input, the first sub-word from pre-trained language models was selected as the input word for the BiLSTM, as detailed in \cite{phobert}. In terms of optimization details, all experiments were conducted over 100 epochs to ensure that all models could converge equally. The batch size was set to 32, and the initial learning rate was 1e-3, which was then decayed using a linear scheduler.

\subsection{Evaluation Metrics}
Drawing upon insights derived from two prior studies focused on Vietnamese span detection, this paper utilizes both a strict evaluation method, as outlined in the work of Thanh et al. (2021) \cite{thanh-etal-2021-span}, and a more permissive evaluation approach, as detailed in the study by Hoang et al. (2023) \cite{hoang-etal-2023-vihos}, to evaluate the performance of predictive models. In the context of the relaxed evaluation, which is inspired by the approach in \cite{hoang-etal-2023-vihos}, we proposed to exclusively assess the positive class. This choice deliberately yields lower metrics and offers ample room for future enhancements and refinements. So, the F-score for the strict evaluation is computed using the following formulations.
\begin{align}
\text{Precision} &= \frac{\text{Total exact matching spans}}{\text{Total spans predicted by the system}} \\
\text{Recall} &= \frac{\text{Total exact matching spans}}{\text{Total gold spans}} \\
\text{F-score} &= \frac{2 \cdot \text{Precision} \cdot \text{Recall}}{\text{Precision} + \text{Recall}}
\end{align}
Similarly, the F-score for the relaxed evaluation is calculated using the following formulas.
\begin{align}
\text{Precision} &= \frac{\text{Total exact matching characters}}{\text{Total characters predicted by the system}} \\
\text{Recall} &= \frac{\text{Total exact matching characters}}{\text{Total gold characters}} \\
\text{F-score} &= \frac{2 \cdot \text{Precision} \cdot \text{Recall}}{\text{Precision} + \text{Recall}}
\end{align}

\subsection{Experiments Results}

%
\begin{table*}[t]
\centering
\caption{Unlabeled abusive span detection performances on dev and test sets}
\label{tab:mainres:1}
\resizebox{\textwidth}{!}{%
\begin{tabular}{|c|c|l|llllll|llllll|}
\hline
\multirow{3}{*}{\rotatebox{90}{\textbf{Decoder~}}} & \multirow{3}{*}{\begin{tabular}[c]{@{}c@{}}\textbf{Training}\\ \textbf
{Strategy}\end{tabular}} & \multicolumn{1}{c|}{\multirow{3}{*}{\begin{tabular}[c]{@{}c@{}} \textbf {Pre-trained}\\  \textbf {Language}\\ \textbf {Model}\end{tabular}}} & \multicolumn{6}{c|}{\textbf{Strict}} & \multicolumn{6}{c|}{\textbf{Relaxed}} \\ \cline{4-15} 
 &  & \multicolumn{1}{c|}{} & \multicolumn{3}{c|}{\textbf{Dev}} & \multicolumn{3}{c|}{\textbf{Test}} & \multicolumn{3}{c|}{\textbf{Dev}} & \multicolumn{3}{c|}{\textbf{Test}} \\ \cline{4-15} 
 &  & \multicolumn{1}{c|}{} & \multicolumn{1}{c|}{\textbf{P}} & \multicolumn{1}{c|}{\textbf{R}} & \multicolumn{1}{c|}{\textbf{F}} & \multicolumn{1}{c|}{\textbf{P}} & \multicolumn{1}{c|}{\textbf{R}} & \multicolumn{1}{c|}{\textbf{F}} & \multicolumn{1}{c|}{\textbf{P}} & \multicolumn{1}{c|}{\textbf{R}} & \multicolumn{1}{c|}{\textbf{F}} & \multicolumn{1}{c|}{\textbf{P}} & \multicolumn{1}{c|}{\textbf{R}} & \multicolumn{1}{c|}{\textbf{F}} \\ \hline
\multirow{8}{*}{\rotatebox{90}{\textbf{Softmax}}} & \multirow{4}{*}{\begin{tabular}[c]{@{}c@{}} \textbf {Unlabeled}\\ \textbf {Span}\\  \textbf {Detection}\end{tabular}} & $\text{XLM-R}_\text{base}$ & \multicolumn{1}{j|}{22.22} & \multicolumn{1}{j|}{24.49} & \multicolumn{1}{j|}{23.30} & \multicolumn{1}{j|}{26.89} & \multicolumn{1}{j|}{27.31} & \multicolumn{1}{j|}{27.10} & \multicolumn{1}{j|}{88.00} & \multicolumn{1}{j|}{80.30} & \multicolumn{1}{j|}{83.97} & \multicolumn{1}{j|}{89.00} & \multicolumn{1}{j|}{82.96} & \multicolumn{1}{j|}{85.88} \\ \cline{3-15} 
 &  & $\text{XLM-R}_\text{large}$ & \multicolumn{1}{j|}{20.49} & \multicolumn{1}{j|}{25.51} & \multicolumn{1}{j|}{22.73} & \multicolumn{1}{j|}{24.54} & \multicolumn{1}{j|}{30.77} & \multicolumn{1}{j|}{27.30} & \multicolumn{1}{j|}{85.98} & \multicolumn{1}{j|}{73.93} & \multicolumn{1}{j|}{79.50} & \multicolumn{1}{j|}{80.54} & \multicolumn{1}{j|}{80.79} & \multicolumn{1}{j|}{80.67} \\ \cline{3-15} 
 &  & $\text{PhoBERT}_\text{base}$ & \multicolumn{1}{j|}{34.69} & \multicolumn{1}{j|}{\textbf{34.69}} & \multicolumn{1}{j|}{\textbf{34.69}} & \multicolumn{1}{j|}{26.39} & \multicolumn{1}{j|}{29.23} & \multicolumn{1}{j|}{27.74} & \multicolumn{1}{j|}{\textbf{95.27}} & \multicolumn{1}{j|}{73.83} & \multicolumn{1}{j|}{83.19} & \multicolumn{1}{j|}{88.27} & \multicolumn{1}{j|}{80.91} & \multicolumn{1}{j|}{84.43} \\ \cline{3-15} 
 &  & $\text{PhoBERT}_\text{large}$ & \multicolumn{1}{j|}{27.78} & \multicolumn{1}{j|}{25.51} & \multicolumn{1}{j|}{26.60} & \multicolumn{1}{j|}{25.90} & \multicolumn{1}{j|}{27.69} & \multicolumn{1}{j|}{26.77} & \multicolumn{1}{j|}{94.59} & \multicolumn{1}{j|}{64.78} & \multicolumn{1}{j|}{76.90} & \multicolumn{1}{j|}{91.35} & \multicolumn{1}{j|}{80.27} & \multicolumn{1}{j|}{85.45} \\ \cline{2-15} 
 & \multirow{4}{*}{\begin{tabular}[c]{@{}c@{}} \textbf {Join Span}\\ \textbf {Detection}\\ \& \textbf {Labeling}\end{tabular}} & $\text{XLM-R}_\text{base}$ & \multicolumn{1}{j|}{10.90} & \multicolumn{1}{j|}{17.35} & \multicolumn{1}{j|}{13.39} & \multicolumn{1}{j|}{7.81} & \multicolumn{1}{j|}{14.12} & \multicolumn{1}{j|}{10.05} & \multicolumn{1}{j|}{91.86} & \multicolumn{1}{j|}{76.15} & \multicolumn{1}{j|}{83.27} & \multicolumn{1}{j|}{91.75} & \multicolumn{1}{j|}{76.72} & \multicolumn{1}{j|}{83.57} \\ \cline{3-15} 
 &  & $\text{XLM-R}_\text{large}$ & \multicolumn{1}{j|}{4.46} & \multicolumn{1}{j|}{10.20} & \multicolumn{1}{j|}{6.21} & \multicolumn{1}{j|}{4.62} & \multicolumn{1}{j|}{11.45} & \multicolumn{1}{j|}{6.58} & \multicolumn{1}{j|}{87.94} & \multicolumn{1}{j|}{74.89} & \multicolumn{1}{j|}{80.89} & \multicolumn{1}{j|}{77.19} & \multicolumn{1}{j|}{82.81} & \multicolumn{1}{j|}{79.90} \\ \cline{3-15} 
 &  & $\text{PhoBERT}_\text{base}$ & \multicolumn{1}{j|}{12.82} & \multicolumn{1}{j|}{20.41} & \multicolumn{1}{j|}{15.75} & \multicolumn{1}{j|}{8.12} & \multicolumn{1}{j|}{16.79} & \multicolumn{1}{j|}{10.95} & \multicolumn{1}{j|}{97.33} & \multicolumn{1}{j|}{70.91} & \multicolumn{1}{j|}{82.04} & \multicolumn{1}{j|}{89.75} & \multicolumn{1}{j|}{78.27} & \multicolumn{1}{j|}{83.62} \\ \cline{3-15}
 &  & $\text{PhoBERT}_\text{large}$ & \multicolumn{1}{j|}{12.88} & \multicolumn{1}{j|}{17.35} & \multicolumn{1}{j|}{14.78} & \multicolumn{1}{j|}{7.71} & \multicolumn{1}{j|}{14.89} & \multicolumn{1}{j|}{10.16} & \multicolumn{1}{j|}{89.66} & \multicolumn{1}{j|}{64.22} & \multicolumn{1}{j|}{74.83} & \multicolumn{1}{j|}{91.49} & \multicolumn{1}{j|}{77.06} & \multicolumn{1}{j|}{83.66} \\ \hline
\multirow{8}{*}{\rotatebox{90}{\textbf{CRF}}} & \multirow{4}{*}{\begin{tabular}[c]{@{}c@{}}\textbf {Unlabeled}\\ \textbf {Span}\\ \textbf {Detection}\end{tabular}} & $\text{XLM-R}_\text{base}$ & \multicolumn{1}{j|}{30.61} & \multicolumn{1}{j|}{30.61} & \multicolumn{1}{j|}{30.61} & \multicolumn{1}{j|}{\textbf{32.54}} & \multicolumn{1}{j|}{31.54} & \multicolumn{1}{j|}{32.03} & \multicolumn{1}{j|}{88.57} & \multicolumn{1}{j|}{\textbf{82.10}} & \multicolumn{1}{j|}{85.21} & \multicolumn{1}{j|}{87.42} & \multicolumn{1}{j|}{82.51} & \multicolumn{1}{j|}{84.90} \\ \cline{3-15} 
 &  & $\text{XLM-R}_\text{large}$ & \multicolumn{1}{j|}{29.41} & \multicolumn{1}{j|}{30.61} & \multicolumn{1}{j|}{30.00} & \multicolumn{1}{j|}{26.21} & \multicolumn{1}{j|}{29.23} & \multicolumn{1}{j|}{27.64} & \multicolumn{1}{j|}{84.23} & \multicolumn{1}{j|}{72.19} & \multicolumn{1}{j|}{77.74} & \multicolumn{1}{j|}{76.81} & \multicolumn{1}{j|}{\textbf{83.12}} & \multicolumn{1}{j|}{79.84} \\ \cline{3-15} 
 &  & $\text{PhoBERT}_\text{base}$ & \multicolumn{1}{j|}{36.36} & \multicolumn{1}{j|}{32.65} & \multicolumn{1}{j|}{34.41} & \multicolumn{1}{j|}{32.33} & \multicolumn{1}{j|}{\textbf{33.08}} & \multicolumn{1}{j|}{\textcolor{red}{\textbf{32.70}}} & \multicolumn{1}{j|}{97.38} & \multicolumn{1}{j|}{73.65} & \multicolumn{1}{j|}{83.87} & \multicolumn{1}{j|}{87.62} & \multicolumn{1}{j|}{82.29} & \multicolumn{1}{j|}{84.87} \\ \cline{3-15} 
 &  & $\text{PhoBERT}_\text{large}$ & \multicolumn{1}{j|}{36.25} & \multicolumn{1}{j|}{29.59} & \multicolumn{1}{j|}{32.58} & \multicolumn{1}{j|}{31.78} & \multicolumn{1}{j|}{28.85} & \multicolumn{1}{j|}{30.24} & \multicolumn{1}{j|}{91.93} & \multicolumn{1}{j|}{68.20} & \multicolumn{1}{j|}{78.31} & \multicolumn{1}{j|}{\textbf{92.72}} & \multicolumn{1}{j|}{80.20} & \multicolumn{1}{j|}{\textcolor{red}{\textbf{86.00}}} \\ \cline{2-15} 
 & \multirow{4}{*}{\begin{tabular}[c]{@{}c@{}} \textbf {Join Span}\\ \textbf {Detection}\\ \& \textbf {Labeling}\end{tabular}} & $\text{XLM-R}_\text{base}$ & \multicolumn{1}{j|}{29.00} & \multicolumn{1}{j|}{29.59} & \multicolumn{1}{j|}{29.29} & \multicolumn{1}{j|}{26.64} & \multicolumn{1}{j|}{27.86} & \multicolumn{1}{j|}{27.24} & \multicolumn{1}{j|}{93.02} & \multicolumn{1}{j|}{78.73} & \multicolumn{1}{j|}{\textbf{85.28}} & \multicolumn{1}{j|}{88.75} & \multicolumn{1}{j|}{80.10} & \multicolumn{1}{j|}{84.20} \\ \cline{3-15} 
 &  & ${\text{XLM-R}}_{\text{large}}$ & \multicolumn{1}{c|}{27.08} & \multicolumn{1}{c|}{26.53} & \multicolumn{1}{c|}{26.80} & \multicolumn{1}{c|}{23.59} & \multicolumn{1}{c|}{25.57} & \multicolumn{1}{c|}{24.54} & \multicolumn{1}{c|}{93.33} & \multicolumn{1}{c|}{67.48} & \multicolumn{1}{c|}{78.33} & \multicolumn{1}{c|}{87.23} & \multicolumn{1}{c|}{77.12} & \multicolumn{1}{c|}{81.86} \\ \cline{3-15} 
 &  & $\text{PhoBERT}_\text{base}$ & \multicolumn{1}{c|}{28.12} & \multicolumn{1}{c|}{27.55} & \multicolumn{1}{c|}{27.84} & \multicolumn{1}{j|}{24.50} & \multicolumn{1}{j|}{27.86} & \multicolumn{1}{j|}{26.07} & \multicolumn{1}{j|}{94.89} & \multicolumn{1}{c|}{69.50} & \multicolumn{1}{c|}{80.24} & \multicolumn{1}{c|}{89.44} & \multicolumn{1}{j|}{80.61} & \multicolumn{1}{j|}{84.80} \\ \cline{3-15}
 &  & $\text{PhoBERT}_\text{large}$ & \multicolumn{1}{c|}{\textbf{37.80}} & \multicolumn{1}{c|}{31.63} & \multicolumn{1}{j|}{34.44} & \multicolumn{1}{j|}{28.40} & \multicolumn{1}{j|}{27.10} & \multicolumn{1}{j|}{27.73} & \multicolumn{1}{j|}{92.63} & \multicolumn{1}{j|}{68.72} & \multicolumn{1}{j|}{78.91} & \multicolumn{1}{j|}{92.40} & \multicolumn{1}{j|}{79.19} & \multicolumn{1}{j|}{85.28} \\ \hline
\end{tabular}
}
\end{table*}

\begin{table*}[t]
\centering
\caption{Labeled abusive span detection F-scores on test set of models using join span detection and labeling training strategy}
\label{tab:mainres:2}
\resizebox{\textwidth}{!}{%
\begin{tabular}{|c|l|ccccccc|ccccccc|}
\hline
\multirow{2}{*}{\rotatebox{90}{\textbf{Decoder~}}} & \multicolumn{1}{c|}{\multirow{2}{*}{\textbf{\begin{tabular}[c]{@{}c@{}}Pre-trained\\ Language\\ Model\end{tabular}}}} & \multicolumn{7}{c|}{\textbf{Strict}} & \multicolumn{7}{c|}{\textbf{Relaxed}} \\ \cline{3-16} 
 & \multicolumn{1}{c|}{} & \multicolumn{1}{c|}{\rotatebox{90}{\textbf{Cultural Identity~}}} & \multicolumn{1}{c|}{\rotatebox{90}{\textbf{Financial Economic~}}} & \multicolumn{1}{c|}{\rotatebox{90}{\textbf{Mental Psychological~}}} & \multicolumn{1}{c|}{\rotatebox{90}{\textbf{Physical~}}} & \multicolumn{1}{c|}{\rotatebox{90}{\textbf{Sexual~}}} & \multicolumn{1}{c|}{\rotatebox{90}{\textbf{Verbal Emotional~}}} & \rotatebox{90}{\textbf{Overall~}} & \multicolumn{1}{c|}{\rotatebox{90}{\textbf{Cultural Identity~}}} & \multicolumn{1}{c|}{\rotatebox{90}{\textbf{Financial Economic~}}} & \multicolumn{1}{c|}{\rotatebox{90}{\textbf{Mental Psychological~}}} & \multicolumn{1}{c|}{\rotatebox{90}{\textbf{Physical~}}} & \multicolumn{1}{c|}{\rotatebox{90}{\textbf{Sexual~}}} & \multicolumn{1}{c|}{\rotatebox{90}{\textbf{Verbal Emotional~}}} & \rotatebox{90}{\textbf{Overall~}} \\ \hline
\multirow{4}{*}{\rotatebox{90}{\textbf{Softmax}}} & $\text{XLM-R}_\text{base}$ & \multicolumn{1}{c|}{0.00} & \multicolumn{1}{c|}{9.09} &  \multicolumn{1}{c|}{5.88} & \multicolumn{1}{c|}{8.62} & \multicolumn{1}{c|}{0.00} & \multicolumn{1}{c|}{12.40} & \multicolumn{1}{c|}{8.15} & \multicolumn{1}{c|}{28.43} & \multicolumn{1}{c|}{\textbf{66.58}} & \multicolumn{1}{c|}{46.88} & \multicolumn{1}{c|}{59.05} & \multicolumn{1}{c|}{0.00} & \multicolumn{1}{c|}{64.55} & \multicolumn{1}{c|}{55.41} \\ \cline{2-16} 
 & $\text{XLM-R}_\text{large}$ & \multicolumn{1}{c|}{0.00} & \multicolumn{1}{c|}{1.96} & \multicolumn{1}{c|}{4.40} & \multicolumn{1}{c|}{0.00} & \multicolumn{1}{c|}{0.00} & \multicolumn{1}{c|}{5.71} & \multicolumn{1}{c|}{4.17} & \multicolumn{1}{c|}{0.00} & \multicolumn{1}{c|}{48.43} & \multicolumn{1}{c|}{42.68} & \multicolumn{1}{c|}{26.71} & \multicolumn{1}{c|}{0.00} & \multicolumn{1}{c|}{51.92} & \multicolumn{1}{c|}{44.43} \\ \cline{2-16} 
 & $\text{PhoBERT}_\text{base}$ & \multicolumn{1}{c|}{0.00} & \multicolumn{1}{c|}{9.26} & \multicolumn{1}{c|}{7.48} & \multicolumn{1}{c|}{5.66} & \multicolumn{1}{c|}{0.00} & \multicolumn{1}{c|}{11.28} & \multicolumn{1}{c|}{8.46} & \multicolumn{1}{c|}{11.89} & \multicolumn{1}{c|}{60.84} & \multicolumn{1}{c|}{55.01} & \multicolumn{1}{c|}{44.66} & \multicolumn{1}{c|}{0.00} & \multicolumn{1}{c|}{63.10} & \multicolumn{1}{c|}{55.70} \\ \cline{2-16} 
 & $\text{PhoBERT}_\text{large}$ & \multicolumn{1}{c|}{0.00} & \multicolumn{1}{c|}{1.79} & \multicolumn{1}{c|}{8.89} & \multicolumn{1}{c|}{2.13} & \multicolumn{1}{c|}{0.00} & \multicolumn{1}{c|}{10.37} & \multicolumn{1}{c|}{7.29} & \multicolumn{1}{c|}{0.00} & \multicolumn{1}{c|}{64.21} & \multicolumn{1}{c|}{54.51} & \multicolumn{1}{c|}{38.10} & \multicolumn{1}{c|}{0.00} & \multicolumn{1}{c|}{61.50} & \multicolumn{1}{c|}{54.55} \\ \hline
\multirow{4}{*}{\rotatebox{90}{\textbf{CRF}}} & $\text{XLM-R}_\text{base}$ & \multicolumn{1}{c|}{0.00} & \multicolumn{1}{c|}{12.90} & \multicolumn{1}{c|}{18.27} & \multicolumn{1}{c|}{8.82} & \multicolumn{1}{c|}{0.00} & \multicolumn{1}{c|}{\textbf{27.06}} & \multicolumn{1}{c|}{18.28} & \multicolumn{1}{c|}{\textbf{38.86}} & \multicolumn{1}{c|}{65.10} & \multicolumn{1}{c|}{51.34} & \multicolumn{1}{c|}{27.12} & \multicolumn{1}{c|}{0.00} & \multicolumn{1}{c|}{\textbf{70.82}} & \multicolumn{1}{c|}{55.41} \\ \cline{2-16} 
 & $\text{XLM-R}_\text{large}$ & \multicolumn{1}{c|}{0.00} & \multicolumn{1}{c|}{\textbf{17.86}} & \multicolumn{1}{c|}{15.38} & \multicolumn{1}{c|}{\textbf{25.00}} & \multicolumn{1}{c|}{0.00} & \multicolumn{1}{c|}{15.05} & \multicolumn{1}{c|}{16.48} & \multicolumn{1}{c|}{0.00} & \multicolumn{1}{c|}{58.52} & \multicolumn{1}{c|}{49.36} & \multicolumn{1}{c|}{\textbf{61.03}} & \multicolumn{1}{c|}{0.00} & \multicolumn{1}{c|}{53.96} & \multicolumn{1}{c|}{51.84} \\ \cline{2-16} 
 & $\text{PhoBERT}_\text{base}$ & \multicolumn{1}{c|}{0.00} & \multicolumn{1}{c|}{15.62} & \multicolumn{1}{c|}{17.53} & \multicolumn{1}{c|}{19.44} & \multicolumn{1}{c|}{0.00} & \multicolumn{1}{c|}{20.59} & \multicolumn{1}{c|}{17.86} & \multicolumn{1}{c|}{26.11} & \multicolumn{1}{c|}{66.50} & \multicolumn{1}{c|}{54.20} & \multicolumn{1}{c|}{58.27} & \multicolumn{1}{c|}{0.00} & \multicolumn{1}{c|}{61.18} & \multicolumn{1}{c|}{57.46} \\ \cline{2-16} 
 & $\text{PhoBERT}_\text{large}$ & \multicolumn{1}{c|}{0.00} & \multicolumn{1}{c|}{17.24} & \multicolumn{1}{c|}{\textbf{18.95}} & \multicolumn{1}{c|}{20.00} & \multicolumn{1}{c|}{0.00} & \multicolumn{1}{c|}{21.95} & \multicolumn{1}{c|}{\textcolor{red}{\textbf{18.75}}} & \multicolumn{1}{c|}{29.24} & \multicolumn{1}{c|}{61.65} & \multicolumn{1}{c|}{\textbf{58.36}} & \multicolumn{1}{c|}{54.11} & \multicolumn{1}{c|}{0.00} & \multicolumn{1}{c|}{64.44} & \multicolumn{1}{c|}{\textcolor{red}{\textbf{58.10}}} \\ \hline
\end{tabular}%
}
\end{table*}

Our study explored two tasks, including unlabeled abusive span detection, and labeled abusive span detection. In the first task, we identified the abusive spans without categorizing them into specific types. In the second task, we not only identified abusive spans but also classified them into six different types of abuse. The results of those tasks were shown in Table \ref{tab:mainres:1}: Experimental results of the first task, and Table \ref{tab:mainres:2}: Experimental results of the second one.

From both tables, we saw that the results of strict evaluation are quite lower than the relaxed method. The reason is that strict evaluation requires precise prediction of the start and end points of a span, while relaxed evaluation allows for more flexibility.

Most experiment results show that using CRF as the final decoder layer provided better predictions than the softmax layer. However, the $\text{XLM-R}_\text{base}$ coupled with BiLSTM-Softmax is an exception. It achieved the best result for Financial/Economic types with a 66.58\% F-score.

For \textit{the unlabeled abusive span detection} experimental results in Table \ref{tab:mainres:1}, on the Test set, $\text{PhoBERT}_\text{base}$ combined with BiLSTM-CRF provided the highest F-score (32.70\%) in the strict evaluation, while $\text{PhoBERT}_\text{large}$ coupled with BiLSTM-CRF outperformed with remains with F-score is 86.00\% in the relaxed evaluation.

For \textit{the labeled abusive span detection} experimental results in Table \ref{tab:mainres:2}, we observed that the Verbal/Emotional type had the highest F-score. Specifically, it achieved a 27.06\% F-score in strict evaluation, and a 70.82\% F-score in relaxed evaluation when using the $\text{XLM-R}_\text{base}$ with BiLSTM-CRF. The distribution of Verbal/Emotional abuse texts within our dataset, which comprised 39.5\%, significantly influenced this outcome. Besides that, the Sexual and Cultural/Identity types had the lowest scores, with a F-score of 0\%. This is because these types comprise only a tiny portion of our dataset, accounting for 2.1\% and 3.1\% respectively. The dataset's imbalance needs to be addressed in the future to enhance predictions for these types of abuse.

We also discovered that using $\text{PhoBERT}_\text{large}$ combined with BiLSTM-CRF produced the highest overall results. This approach achieved the highest F-scores in both evaluation methods, with an F-score of 18.75\% for strict evaluation and 58.10\% for relaxed evaluation.

Although the F-scores in strict evaluation were relatively low, it indicates that there is still room for improvement in the future. On the other hand, the promising results in the relaxed evaluation suggest a potential for further development in detecting abusive spans in Vietnamese narrative texts.


\subsection{Error Analysis}
\begin{figure}[t]
    \centering
    \includegraphics[width=\columnwidth]{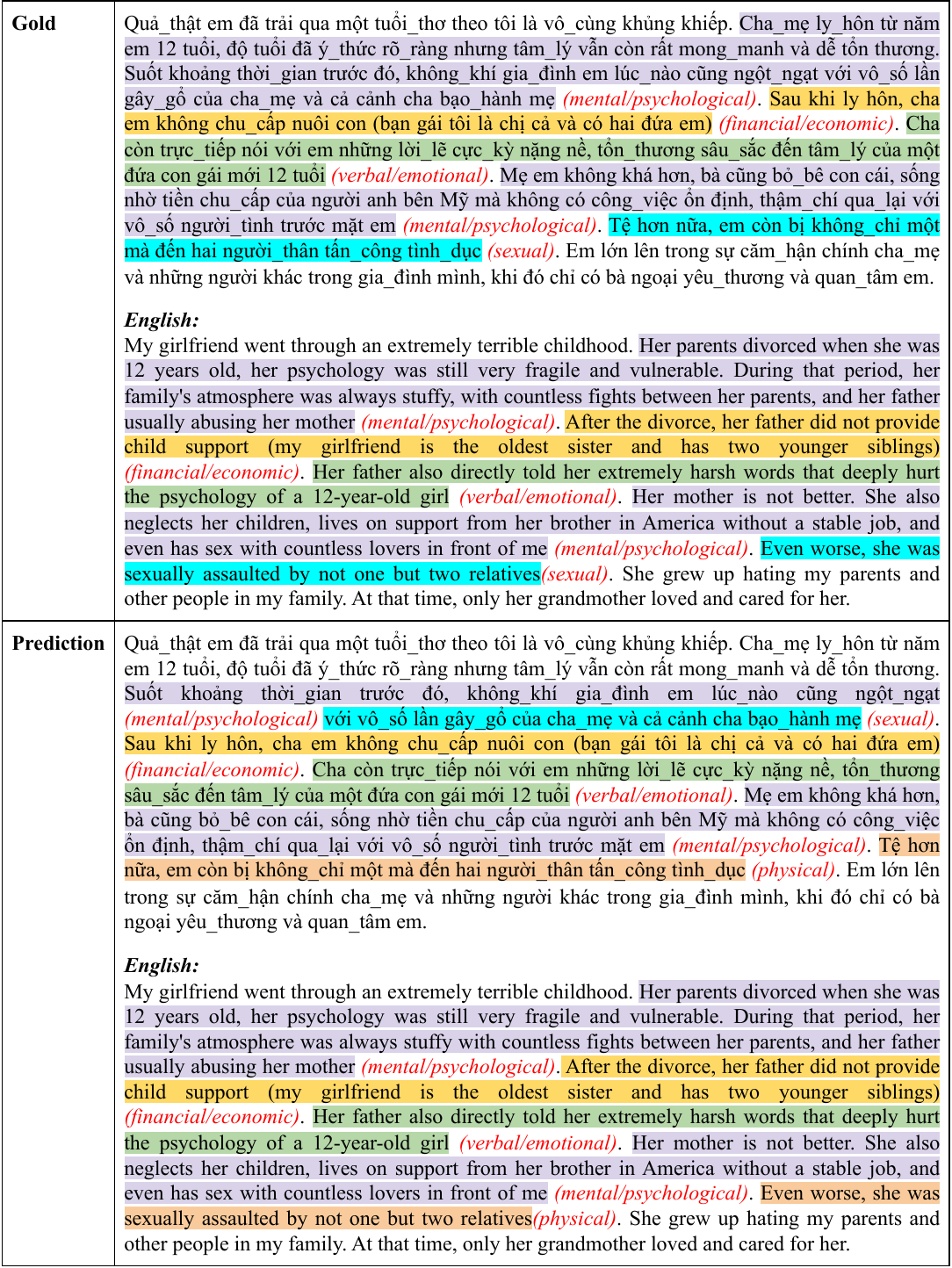}
    \caption{A prediction from $\text{PhoBERT}_\text{large}$ for labeled abusive span detection}
    \label{fig:labeled-phobertlarge}
    \vspace{-2.5mm}
\end{figure}

We visualized a prediction of the labeled abusive span detection task in the Test set of $\text{PhoBERT}_\text{large}$ coupled with BiLSTM-CRF. The results are shown in Figure \ref{fig:labeled-phobertlarge}. We observed that the Verbal/Emotional, Financial/Economic, and Mental/Psychological abuses were accurately detected with precise start and end points. However, the Sexual abuse (span in blue color) was incorrectly predicted as Physical (span in orange color) abuse, and the span of Mental/Psychological abuse (span in purple color) was mistakenly predicted as Sexual abuse. The reason is that the limited training data for the Sexual abuse, and future improvements are necessary to achieve more accurate identification.



\section{Conclusion and Future Works}
We presented the first Vietnamese benchmark dataset for the new task namely Abusive spans detection in Vietnamese narrative texts. The dataset can be served with two tasks, including the unlabeled abusive span detection, and the labeled abusive span detection. For the labeled abusive span detection, the dataset consists of 519 spans in 198 abusive texts in 1041 narrative texts annotated with six aspect categories: Physical, Sexual, Financial/Economic, Cultural/Identity, Verbal/Emotional, and Mental/Psychological. We also conducted experiments with models XLM-RoBERTa, PhoBERT coupled with BiLSTM, and utilized two kinds of decoders: Softmax and CRF for the classifier layer. As a result, we observed that using CRF as the final decoder layer provides better predictions than the Softmax layer. Most importantly, using {$\text{PhoBERT}_\text{large}$}  combined with BiLSTM-CRF outperformed other models with the highest overall results, at 86.00\% and 58.10\% respectively for the relaxed evaluation for both two tasks.

To enhance our results, we must address several limitations in the future. Firstly, our dataset is imbalanced among abusive types, notably in the Sexual and Cultural/Identity categories. To improve predictive accuracy, we will gather more data for these categories. Secondly, our current system can only predict a single abusive type for each span, whereas real-world abusive spans may encompass multiple abusive types simultaneously. Additionally, we have not yet accounted for detecting nested spans in this study, an area for improvement that should be considered in the future.

\begin{acks}
This research was supported by The VNUHCM-University of Information Technology’s Scientific Research Support Fund. We are grateful to the reviewers for their valuable comments, which have greatly enhanced the quality of our work.
\end{acks}

\balance
\bibliographystyle{ACM-Reference-Format}
\bibliography{main}

\end{document}